\documentclass[journal]{IEEEtran}
\ifCLASSINFOpdf
\else
\fi
\usepackage{amsmath}
\usepackage{graphicx}
\usepackage{mdwtab}
\usepackage{booktabs}
\usepackage{multirow}
\usepackage{multicol}
\usepackage{xcolor}
\usepackage{amssymb}
\usepackage{hyperref}
\usepackage[short]{optidef}
\usepackage[ruled,vlined]{algorithm2e}

\newcommand\Y{\mathbf Y}
\newcommand\E{\mathbf E}
\newcommand\A{\mathbf A}

\newcommand\D{\mathbf D}
\newcommand\X{\mathbf X}
\newcommand\N{\mathbf N}
\newcommand\B{\mathbf B}

\newcommand\1{\bf 1}

\newcommand\abold{\mathbf a}
\newcommand\bbold{\mathbf b}

\newcommand\xbf{\mathbf x}

\newcommand\ie{{\it{i.e. }}}

\begin{document}
%
\title{SUnAA: Sparse Unmixing using Archetypal Analysis}
%
%
%
\author{Behnood~Rasti*,~\IEEEmembership{Senior~Member,~IEEE,}
Alexandre~Zouaoui*,~\IEEEmembership{Student Member,~IEEE,} Julien~Mairal,~\IEEEmembership{Senior Member,~IEEE} and  Jocelyn Chanussot, ~\IEEEmembership{Fellow~Member,~IEEE} 
 \thanks{Behnood Rasti (corresponding author) is with Helmholtz-Zentrum Dresden-Rossendorf, Helmholtz Institute Freiberg for Resource Technology, Machine Learning Group, Chemnitzer Straße 40, 09599 Freiberg, Germany; b.rasti@hzdr.de, behnood.rasti@gmail.com}
\thanks{Alexandre Zouaoui, Jocelyn Chanussot, and Julien Mairal are with Univ. Grenoble Alpes, Inria, CNRS, Grenoble INP, LJK, 38000 Grenoble, France}
\thanks{* Behnood Rasti and Alexandre Zouaoui had an equal contribution.}
\thanks{Manuscript received .....}}

\maketitle

\begin{abstract}
This paper introduces a new sparse unmixing technique using archetypal analysis (SUnAA). First, we design a new model based on archetypal analysis. We assume that the endmembers of interest are a convex combination of endmembers provided by a spectral library and that the number of endmembers of interest is known. Then, we propose a minimization problem. Unlike most conventional sparse unmixing methods, here the minimization problem is non-convex. We minimize the optimization objective iteratively using an active set algorithm. Our method is robust to the initialization and only requires the number of endmembers of interest. SUnAA is evaluated using two simulated datasets for which results confirm its better performance over other conventional and advanced techniques in terms of signal-to-reconstruction error. SUnAA is also applied to Cuprite dataset and the results are compared visually with the available geological map provided for this dataset. The qualitative assessment demonstrates the successful estimation of the minerals abundances and significantly improves the detection of dominant minerals compared to the conventional regression-based sparse unmixing methods. The Python implementation of SUnAA can be found at: https://github.com/BehnoodRasti/SUnAA.
\end{abstract}


%
\IEEEpeerreviewmaketitle

\section{Introduction}
%
%
%
%

\IEEEPARstart{S}{pectral} unmixing estimates the abundances of pure spectra of materials called endmembers. Depending on the prior knowledge available about endmembers, the unmixing problem can be divided into three main categories: (1) Supervised Unmixing, (2) Blind Unmixing and (3) Semi-supervised or Sparse Unmixing. In supervised unmixing, abundances are estimated by relying on known endmembers whereas blind unmixing estimates both the endmembers and the abundances simultaneously. Semi-supervised unmixing relies on a library of endmembers that ideally contains the endmembers present in the scene and is often formulated as a sparse regression problem, thus it is known as sparse unmixing. Abundances can typically be estimated by enforcing sparsity-promoting penalties. J. M. Bioucas-Dias originally proposed this idea in \cite{SUnSAL} where sparse unmixing by variable splitting and augmented Lagrangian (SUnSAL), and the constrained SUnSAL (C-SUnSAL) were introduced. Both SUnSAL and C-SUnSAL use the $\ell_1$ penalty to promote sparsity on the abundances. SUnSAL assumes the $\ell_2$ norm for the fidelity term augmented to the $\ell_1$, while C-SUnSAL assumes the $\ell_2$ norm as a constraint to minimize $\ell_1$. SUnSAL solves the minimization problems using the alternating direction method of multipliers (ADMM) \cite{ADMM}. 

SUnSAL was later improved in \cite{SUnSAL-TV} by adding the total variation (TV) penalty (SUnSAL-TV) to the minimization problem in order to incorporate spatial information. We should note that SUnSAL-TV does not hold the abundances sum-to-one constraint (ASC) due to the conflict with the $\ell_1$ penalty. Collaborative sparse \cite{CoSUn} unmixing enforces the sum of $\ell_2$ norms on the abundances to promote sparsity. Double reweighted sparse unmixing \cite{DRSU} and spectral–spatial weighted sparse unmixing (S$^2$WSU) \cite{SZhang2018} exploit the weighted $\ell_1$ norm to promote sparsity. The former also uses the TV penalty to capture spatial information. Multiscale sparse unmixing algorithm (MUA) \cite{RABorsoi2019} captures spatial correlations by performing sparse regression on segmented pixels using either a binary partition tree (BPT), the simple linear iterative clustering (SLIC), or the K-means algorithm. In \cite{TanerInce2020}, SLIC was chosen for the segmentation, and sparse unmixing was performed using superpixel-based graph Laplacian regularization. \textcolor{black}{ A library pruning-based sparse unmixing called multiple signal classification collaborative sparse unmixing (MUSIC-CSR) was proposed in \cite{MUSIC_CSR}. The library was pruned using an orthogonal projection where HySime \cite{HySime} was used to obtain the subspace bases to reduce the noise effect. Then collaborative sparse regression was used for abundance estimation.}

A common drawback of the sparse unmixing techniques mentioned above is that the estimated fractional abundances do not necessarily describe the aerial fraction of each pure material on the ground due to the absence of ASC. Indeed, the $\ell_1$ penalties cannot be applied to the abundances while holding the ASC. This issue was addressed in sparse unmixing using convolutional neural network (SUnCNN) \cite{SUnCNN}. In \cite{SUnCNN}, it was shown that the problem of selecting a suitable prior for a sparse regression could be moved to the optimization on the parameters of a deep encoder-decoder network while the ASC can be enforced using a softmax layer. However, selecting suitable hyperparameters for deep networks is often challenging. In \cite{SMALU}, an asymmetric encoder-decoder architecture is used for sparse unmixing. Instead of softmax, a sparse variation of softmax is used to avoid the full support of softmax while enforcing ASC. 

In conventional sparse unmixing, the endmembers library is fixed, and the abundances estimation is of interest. However, even a pruned and well-selected spectral library cannot flawlessly represent the endmembers of materials in a real-world dataset. There are several factors, such as noise, atmospheric effects, illumination variations, and the intrinsic variation of materials which may affect the endmembers and induce scaling factors for the endmembers present in the scene compared to the ones from the library. To address this issue, we assume that endmembers of interest can be modeled by a convex combination of the library endmembers. This corresponds to the formulation of archetypal analysis (AA) \cite{cutler1994archetypal}. Recently, archetypal analysis  has been successfully harnessed for blind unmixing in \cite{EDAA}. In \cite{L1AA}, $\ell_1$ sparsity-constrained archetypal analysis was proposed for blind unmixing where the sparsity was enforced on the abundances. In this paper, we propose solving sparse unmixing using archetypal analysis (SUnAA). In the proposed model, an additional matrix is introduced, which defines the contributions of the endmembers from the library to the estimated spectra of endmembers present in the scene. Here the ASC can be enforced but the resulting proposed minimization is jointly non-convex. The optimization problem is solved using an active set algorithm, leading to a parameter-free technique, besides the number of endmembers of interest that is required. The experimental results confirm that SUnAA outperforms conventional and deep learning-based sparse unmixing techniques in terms of signal-to-reconstruction error (SRE) for two simulated datasets and visually for the Cuprite dataset. The major contributions of this paper are summarized as follows: 1) we propose a new model based on archetypal analysis for sparse unmixing: we assume that the unknown endmembers are a convex combination of the library endmembers, 2) we propose a non-convex optimization for sparse unmixing: unlike the conventional sparse unmixing which is based on sparse regression and convex optimization, we show that the proposed non-convex optimization leads to accurate abundances estimation, and 3) we adopt a parameter-free active set algorithm to minimize the proposed optimization problem.


\section{Methodology}
\label{sec: model}

\subsection{Conventional Sparse Unmixing}
In conventional sparse unmixing, the observed spectra are modeled as a linear combination of the library endmembers:
\begin{equation} \label{eq: M1}
{\Y} = {\D} {\X} + {\N},
\end{equation}
where $\Y \in \mathbb{R}^{p\times n}$  denotes the observed spectra over $p$ channels and $\N \in \mathbb{R}^{p\times n}$ is the model error and noise. $\D  \in \mathbb{R}^{p\times m} $ denotes the spectral library containing $m$ endmembers. $\X \in \mathbb{R}^{m\times n}$ is the unknown, potentially redundant, fractional abundances to estimate. 

In sparse unmixing, the (redundant) fractional abundances $\X$  are estimated by applying sparsity-enforcing penalties/constraints in a sparse regression formulation such as:
\begin{argmini}
  {\X}{\frac{1}{2}\|\Y - \D \X\|_F^2 + \lambda \sum_{i=1}^n||{\xbf_i}||_q ,}{\label{eq: PLSU}}{}
  \addConstraint{\X}{ \geq 0}
  \addConstraint{{\bf \1}_m^T \X}{= {\1}_n^T}
\end{argmini}
where $\X = [\xbf_1, \ldots, \xbf_m ]$. 
The $\ell_q$ norm is often selected to be a (weighted) sparsity promoting norm. 
For instance, SUnSAL \cite{SUnSAL_C-SUnSAL} solves problem (\ref{eq: PLSU}) for $q=1$ using ADMM \cite{ADMM}. 
However, \cite{SUnSAL-TV} suggests using SUnSAL without ASC due to the conflict with $\ell_1$.


\subsection{SUnAA}
Inspired by archetypal analysis \cite{cutler1994archetypal}, we propose a new model formulation for sparse unmixing, provided that the number of endmembers of interest, $r$, is known:
\begin{argmini!}
  {\B, \A}{\|\Y - \D \B \A\|_F^2,}{\label{eq: PLSU2}}{(\hat{\B}, \hat{\A}) = }
  \addConstraint{\B}{ \geq 0 \label{eq: ENC}}
  \addConstraint{{\bf \1}_m^T \B}{= {\1}_r^T \label{eq: ESC}}
  \addConstraint{\A}{ \geq 0 \label{eq: ANC}}
  \addConstraint{{\bf \1}_r^T \A}{= {\1}_n^T \label{eq: ASC}}
\end{argmini!}
where $\B \in \mathbb{R}^{m\times r}$ corresponds to the contributions of the endmembers from the library $\D$ and $\A \in \mathbb{R}^{r\times n}$ is the (low-rank) abundance matrix. \textcolor{black}{ We should note that AA uses $\Y$ instead of $\D$ in  (\ref{eq: ENC}). It is worth mentioning that $\D$ turns the blind scenario into a semi-supervised scenario, which could be more efficient in the highly mixed scenario without pure pixels. }

In (\ref{eq: PLSU2}), we assume that the unknown endmembers of interest are a convex combination of the library endmembers. \textcolor{black}{This is our primary assumption to compensate for the library mismatch.} In addition, we assume that the number of endmembers is known. Therefore, we enforce non-negativity and sum-to-one constraints on $\B$. Equivalently to abundance non-negativity constraint (ANC \ref{eq: ANC}) and abundance sum-to-one constraint (ASC \ref{eq: ASC}), we call the constraints on $\B$ endmember non-negativity constraint (ENC \ref{eq: ENC}) and endmember sum-to-one constraint (ESC \ref{eq: ESC}). 

It is important to note that the minimization problem (\ref{eq: PLSU2}) is not jointly convex in $(\B, \A)$. However it is convex with respect to one of the variables  when the other is fixed, hence (\ref{eq: PLSU2}) can be solved by alternating between two steps inside a cyclic descent scheme. First, the $\A$-step when $\B$ is fixed. Second, the $\B$-step, by fixing $\A$. In this paper, we adopt the algorithm proposed in \cite{RAA}. Here, we explain briefly solutions to the sub-problems proposed. 
   $\A$-step: Assuming $\B$ is fixed and $\E = \D \B$, the sub-problem corresponds to
\begin{argmini}
  {\A}{\|\Y - \E \A\|_F^2.}{\label{eq: Astep}}{\hat{\A} = }
  \addConstraint{\A}{ \geq 0}
  \addConstraint{{\bf \1}_r^T \A}{= {\1}_n^T}
\end{argmini}
(\ref{eq: Astep})  is a smooth least-squares optimization problem with a simplicial constraint. As noted by \cite{RAA}, generic quadratic programming solvers could be used but significantly faster convergence can be obtained by designing a dedicated algorithm that can leverage the underlying sparsity of the solution. Following \cite{RAA} we use an active set algorithm to solve (\ref{eq: Astep}). 
      $\B$-step: Assuming  $\A$ is fixed, problem (\ref{eq: PLSU2}) writes as follows:
\begin{argmini}
  {\B}{\|\Y - \D \B \A\|_F^2.}{\label{eq: Bstep}}{\hat{\B} = }
  \addConstraint{\B}{ \geq 0}
  \addConstraint{{\1}_m^T \B}{= {\1}_r^T}
\end{argmini}
Solving (\ref{eq: Bstep}) is not as straightforward as (\ref{eq: Astep}) since it does not correspond to the standard quadratic form.
However, following \cite{RAA}, we consider solving (\ref{eq: Bstep}) separately for every column $\bbold_j$ of $\B = [\bbold_1, \ldots, \bbold_r]$ by fixing all other variables in order to obtain a quadratic program:
\begin{argmini}
  {\bbold_j}{\left|\left|\frac{1}{\|\abold^j\|_2^2} (\Y - \D \B_{\text{old}} \A) \abold^{j \top} + \D \bbold_{j, \text{old}} - \D \bbold_j \right|\right|_F^2,}{\label{eq: Bsubstep}}{}
  \addConstraint{\bbold_j}{ \geq 0}
  \addConstraint{{\1}_m^T \bbold_j}{= 1}
\end{argmini}
where $\bbold_{j, \text{old}}$ is the current value of $\bbold_j$ before the update, and $\abold^{j}$ in $\mathbb{R}^{1 \times n}$ is the $j$-th row of $\A$.

Note that the first and second terms are fixed, therefore the same active set algorithm can be used to solve (\ref{eq: Bsubstep}). The pseudo-code for the algorithm used to solve (\ref{eq: PLSU2}) is given in Algorithm~\ref{Alg:SUnAA}. As for initialization, we uniformly initialize matrices $\B$ and $\A$, \ie $\B^{(0)}=({\1}_{m}{\1}_{r}^{T})/m$ and $\A^{(0)} = ({\1}_{r}{\1}_{n}^{T})/r$.
\begin{algorithm}[tbp]\footnotesize
\SetAlgoLined
\vspace{.cm}
\KwIn{${\bf Y}$: Hyperspectral data, ${\bf D}$: Endmember library, $r$: Number of endmembers, $T$: Number of iterations.}
\vspace{.cm}
\KwOut{$\hat{\A}$: Abundances, $\hat{\E}$: Endmembers, $\hat{\B}$: Endmembers contributions.}
\textbf{Initialization}:  $\B^{(0)}=({\1}_{m}{\1}_{r}^{T})/m$ and $\A^{(0)} = ({\1}_{r}{\1}_{n}^{T})/r$\\
\For{$t = 1$ \KwTo $T$}{
\textbf{A-step}:\\ 
  $\E = \D \B$\\
  $\hat{\A} =\mbox{ActSet}(\Y,\E)$\\
\textbf{B-step}:\\ 
\For{$j=1$ \KwTo $r$}{$
\Tilde{\Y} = \frac{1}{\|\abold^j\|_2^2} (\Y - \D \B_{\text{old}} \A) \abold^{j \top} + \D \bbold_{j, \text{old}}$\\
$\hat{\bbold}_j = \mbox{ActSet}(\Tilde{\Y}, \D)$
}
}
$\hat{\E} = \D \hat{\B}$
\caption{{SUnAA}}
\label{Alg:SUnAA}
\end{algorithm}

\section{Experimental Results}
\label{sec: ex}
We compare SUnAA with seven sparse unmixing techniques: SUnSAL \cite{SUnSAL_C-SUnSAL}, SUnSAL-S (SUnSAL with ASC and without sparsity penalty, \ie $\lambda=0$), SUnSAL-TV \cite{SUnSAL-TV}, S$^2$WSU \cite{SZhang2018}, MUA (using BPT segmentation) \cite{RABorsoi2019}, \textcolor{black}{MUSIC-CSR \cite{MUSIC_CSR},} and SUnCNN \cite{SUnCNN} applied to two simulated datasets and Cuprite. All the parameters are set as default for the competing methods. The results are mean values over ten experiments. 

\subsection{Simulated Datasets}
Two synthetic datacubes (DC1 and DC2) were used for simulated experiments. For DC1, a synthetic library composed of 240 spectral signatures selected from the USGS library with a minimum pair-spectra angle of 4.44\textdegree is used. DC1 was simulated using a linear mixing model with 5 endmembers selected from the library and 75$\times$75 pixels. The abundance maps are composed of five rows of square regions uniformly distributed over the spatial dimension. DC2 is a challenging simulated dataset with no pure pixels and two mixed pixels on the facet of the data simplex (more details in \cite{MiSiCNet}). It contains  105$\times$105 pixels, simulated by the linear combination of six endmembers from the USGS library. Two endmembers show no absorption features throughout the wavelength and are similar but scaled versions of each other, making the scenario even more challenging. 

For the quantitative evaluation, we use the SRE in dB 
\begin{equation}
    \text{SRE}(\X, \hat{\X})=20\log_{10}\frac{\|{\X}\|_F}{\|{\X}-\hat{\X}\|_F}.
\end{equation}
Three levels of additive noise, \ie 20, 30, and 40 dB are considered in the simulated experiments.

Tables \ref{tab:SRE_DC1} and \ref{tab:SRE_DC2} compare the results obtained by applying different sparse unmixing techniques to DC1 and DC2, respectively, in terms of SRE. For DC1, SUnAA outperforms the other techniques for SNR=30 and 40 dB, and for SNR=20 dB gives the second-best results after MUA. SUnCNN shows the second-best performance in terms of SRE. MUA shows the best performance for low SNR, \ie 20 dB. However, it performs poorly for high SNRs. SUNSAL and SUnSAL-S yield very low SREs. S$^2$WSU, MUSIC-CSR and SUnSAL-TV perform moderately, however, S$^2$WSU provides better performances for higher SNRs. 

The results for DC2 show that SUnAA outperforms the other techniques for all SNRs considerably. All the other techniques give low SRE. This could be attributed to the complexity of the dataset. Since the library may have several scaled versions of one endmember, the conventional sparse regression might not lead to a sparse solution. A remedy to this problem is to prune the library by removing the spectra with small spectral angle distances. However, in the case  of complex datasets such as DC2, the endmembers may be removed from the library, which also leads to poor estimation. 




Overall, comparing the SRE reveals that SUnAA outperforms the other techniques. SUnAA also show consistent performances with respect to the noise level and datasets. All the other methods fail for the complex dataset. For the simple datasets, SUnCNN gives the second-best performance. MUA outperforms the other techniques for very low SNR, \ie 20 dB, which could be attributed to the segmentation applied before the sparse regression in MUA. However, this might cause oversmoothing in the abundances. S$^2$W performs moderately and better than SUnSAL models. SUnSAL-TV outperforms SUnSAL and SUnSAL-S due to the TV penalty, which exploits spatial information. 

\begin{figure*} [htbp]\begin{center}
\begin{tabular}{c} 
\includegraphics[width=.6\linewidth]{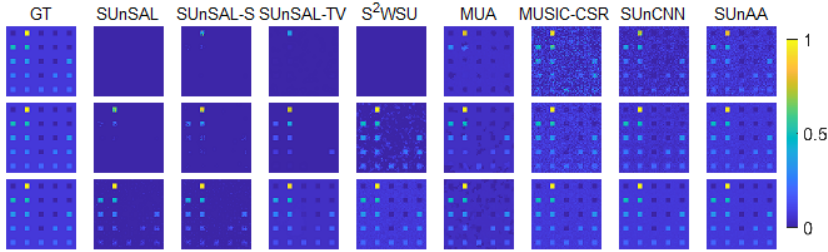}
 \end{tabular} \end{center} \caption{\textcolor{black}{The fractional abundance of endmember 1 of DC1. From top to bottom SNR= 20, 30, and 40 dB.}}
 \label{DC1}
\end{figure*}

\begin{figure*} [htbp]\begin{center}
\begin{tabular}{c} 
\includegraphics[width=.6\linewidth]{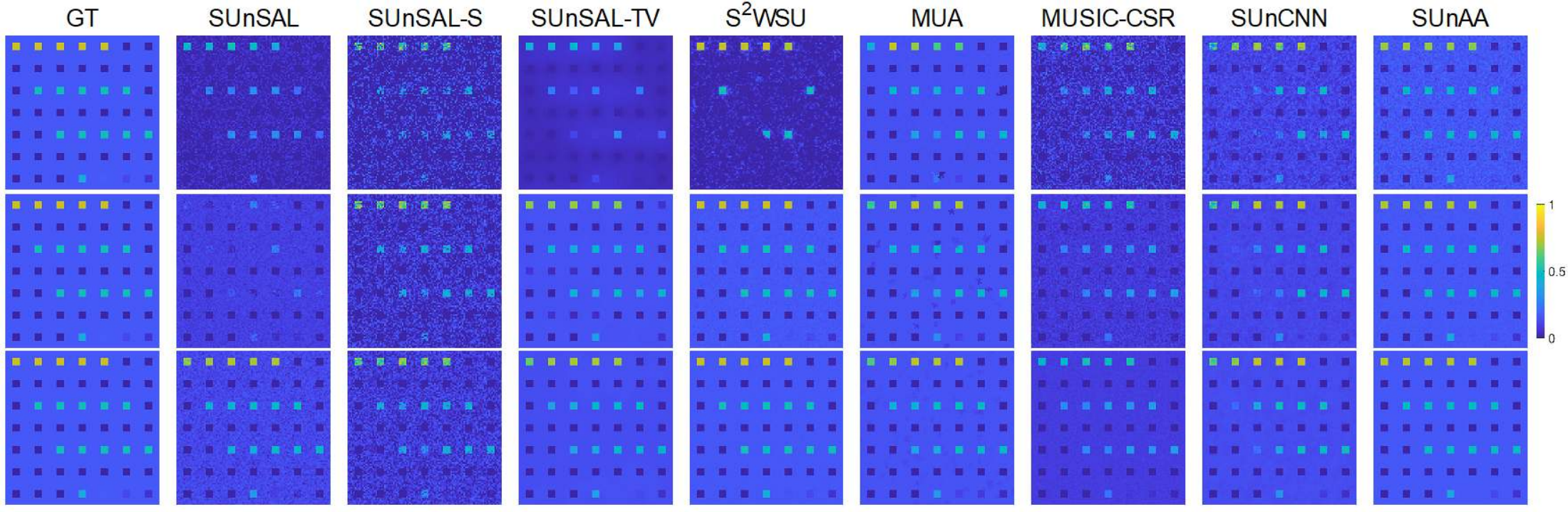}\\
 \end{tabular} \end{center} \caption{\textcolor{black}{The fractional abundance of endmember 1 of DC2. From top to bottom SNR= 20, 30, and 40 dB.}}
 \label{DC2}
\end{figure*}

Fig. \ref{DC1} and \ref{DC2} provide a visual comparison of the abundance maps estimated for endmember 1 of DC1 and DC2, respectively. As can be seen, both MUA and SUnSAL-TV oversmooth the abundances, particularly for low-SNRs. MUA also introduces artifacts, which is not desirable. This is due to the segmentation step in MUA and the absence of parameter selection for the TV regularizer in SUnSAL-TV.

\begin{table}[htbp]\scriptsize
\centering\addtolength{\tabcolsep}{-5.5pt}
\textcolor{black}{	\caption{Sparse unmixing experiments applied to DC1 in terms of SRE. The best performances are shown in bold. The second best are underlined.}
\begin{tabular}{c|c|c|c|c|c|c|c|c}
		\toprule
		SNR    &SUnSAL   & SUnSAL-S & SUnSAL-TV             &S$^2$WSU & MUA & MUSIC-CSR & SUnCNN & SUnAA \\		
		\midrule
		20 dB                &  4.86 & 4.31 & 9.76 & 7.99 & \textbf{13.19} & 6.07& 11.15 & \underline{11.52}  \\
        30 dB                & 8.94 & 8.47 & 14.39 & 15.52 & 18.28 & 13.36& \underline{20.63}   & \textbf{21.27} \\
        40 dB                & 13.83 & 13.15 & 20.84 & 28.16 & 21.12 &24.39 & \underline{30.62} & \textbf{31.23} \\
		\bottomrule
	\end{tabular}        
	\label{tab:SRE_DC1}}
\end{table}

\begin{table}[htbp]\scriptsize
\centering\addtolength{\tabcolsep}{-5.5pt}
\textcolor{black}{	\caption{Sparse unmixing experiments applied to DC2 in terms of SRE. The best performances are shown in bold. The second best are underlined.}
\begin{tabular}{c|c|c|c|c|c|c|c|c}
		\toprule
		SNR    &SUnSAL   & SUnSAL-S & SUnSAL-TV             &S$^2$WSU & MUA & MUSIC-CSR & SUnCNN & SUnAA \\		
		\midrule
		20 dB                & 3.04 & 1.41 & 2.49 & 2.76 & \underline{6.95} &2.51 & 4.45 &  \textbf{9.54} \\
        30 dB                & 3.72 & 2.42 & \underline{7.42} & 6.57 & 6.90 & 4.43& 5.08 & \textbf{10.76} \\
        40 dB                & 5.96 & 3.74 & 6.76 & \underline{7.43} & 7.31 & 5.45& 5.96 & \textbf{11.74} \\
		\bottomrule
	\end{tabular}        
	\label{tab:SRE_DC2}}
\end{table}

\subsection{Cuprite dataset}

\begin{figure*} [htbp]
\centering
\begin{tabular}{cc} 
\includegraphics[width=.14\linewidth]{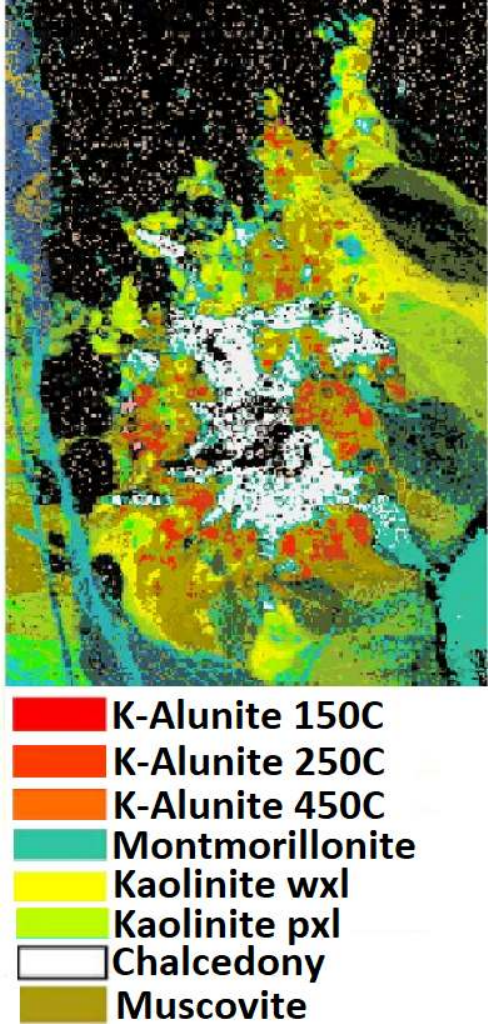}&
\includegraphics[width=.6\linewidth]{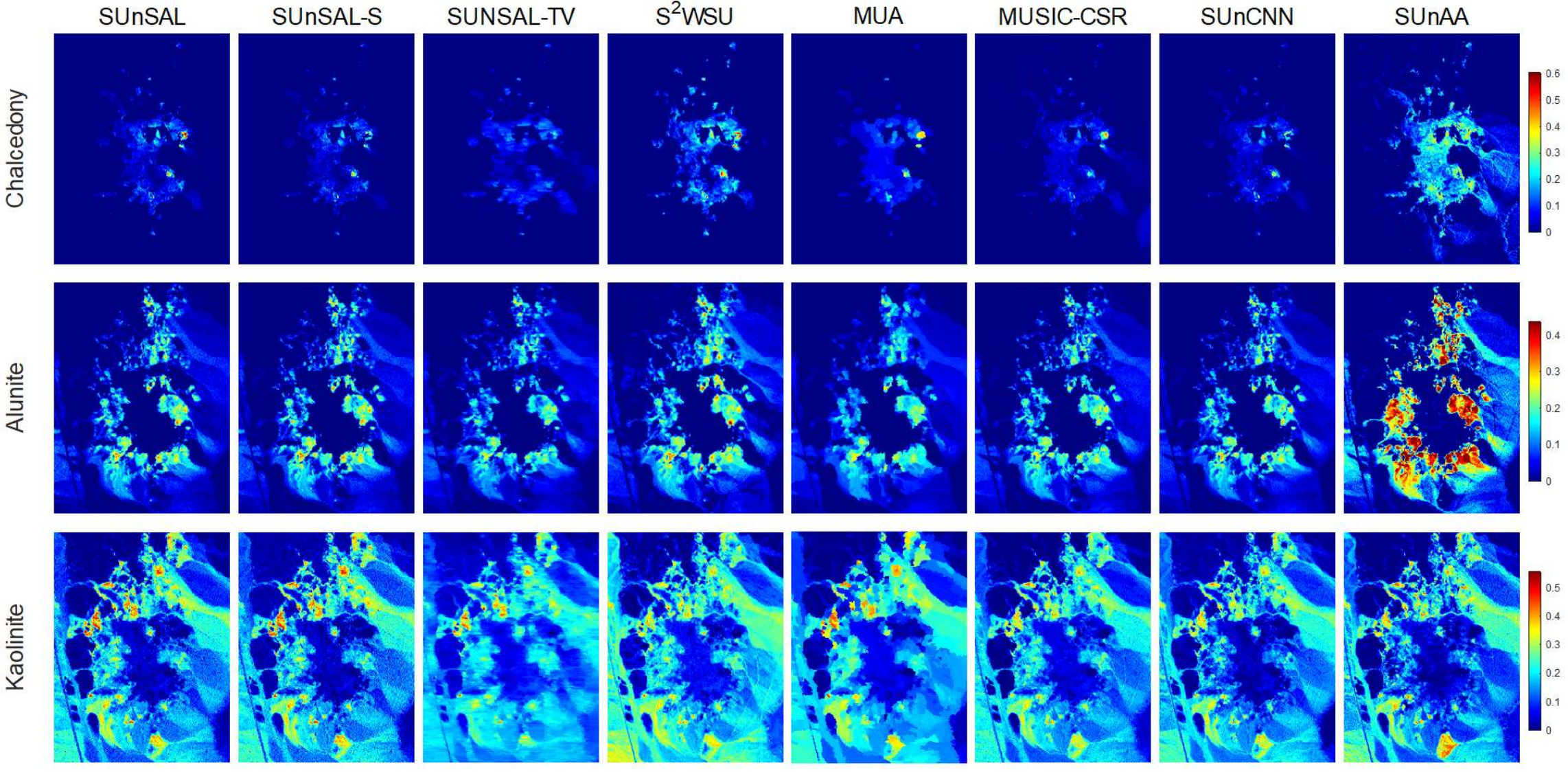}\\
 (a) Geological Map & (b) Estimated abundance maps
 \end{tabular} 
\caption{\textcolor{black}{Abundance maps of three dominant minerals estimated using different sparse unmixing techniques applied to the Cuprite dataset.}}
\label{Real dataset}
\end{figure*}

We used a subset of 250$\times$191 pixels of the Cuprite dataset for real-world experiments. The minerals in that region are well-studied and are therefore suitable for evaluating the abundance maps qualitatively. Fig. \ref{Real dataset} (a) depicts the geological ground reference for the dominant minerals. We use a library $\D \in \mathbb{R}^{188\times 498}$, which is composed of 498 spectral pixels from the USGS library. The water absorption and noisy bands were removed, hence the final pixels are of dimension $p=188$.

Fig. \ref{Real dataset} (b) demonstrates the abundance maps estimated by using different unmixing techniques applied to Cuprite. We showed three dominant minerals in the scene, \ie Chalcedony, Alunite, and Kaolinite, corresponding to library endmembers 297, 420 and 465. For SUnAA, those abundances appear as 15, 13, and 11th. Note that we select $r=16$.  

Fig. \ref{Real dataset} (b) shows that all conventional sparse unmixing techniques and SUnCNN perform similarly. It can be observed that SUnSAL-TV and MUA oversmooth the mineral abundances, which could be attributed to the total variation penalty in SUnSAL-TV and the segmentation-based framework in MUA, which cannot preserve the textures. On the other hand, SUnAA provides sharper maps. Compared to the geological map of USGS (Fig. \ref{Real dataset} (a)), SUnAA considerably outperforms the other techniques for the detection of Chalcedony and Alunite. In the case of Kaolinite, all techniques perform similarly, however, SUnAA shows slightly better performance compared to the other techniques, particularly for the southern region. 

\begin{figure} [htbp]
\centering
\begin{tabular}{cc} 
\includegraphics[width=.4\linewidth]{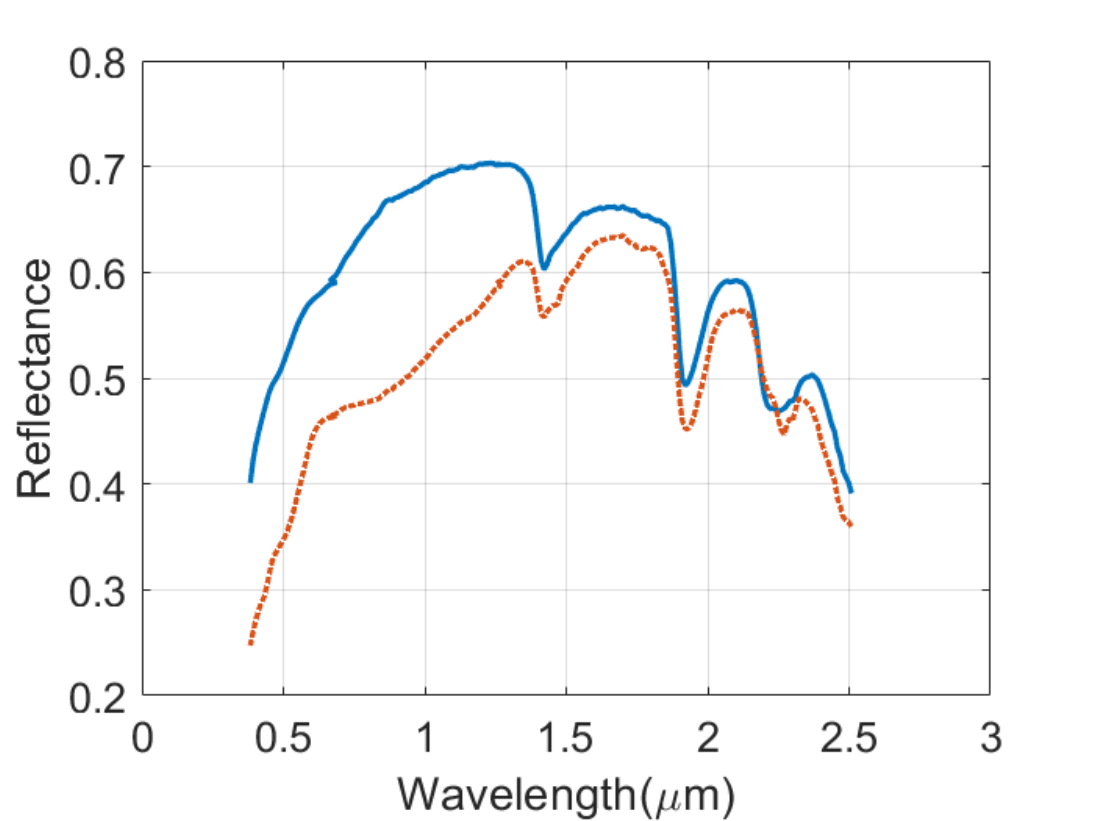}&
\includegraphics[width=.4\linewidth]{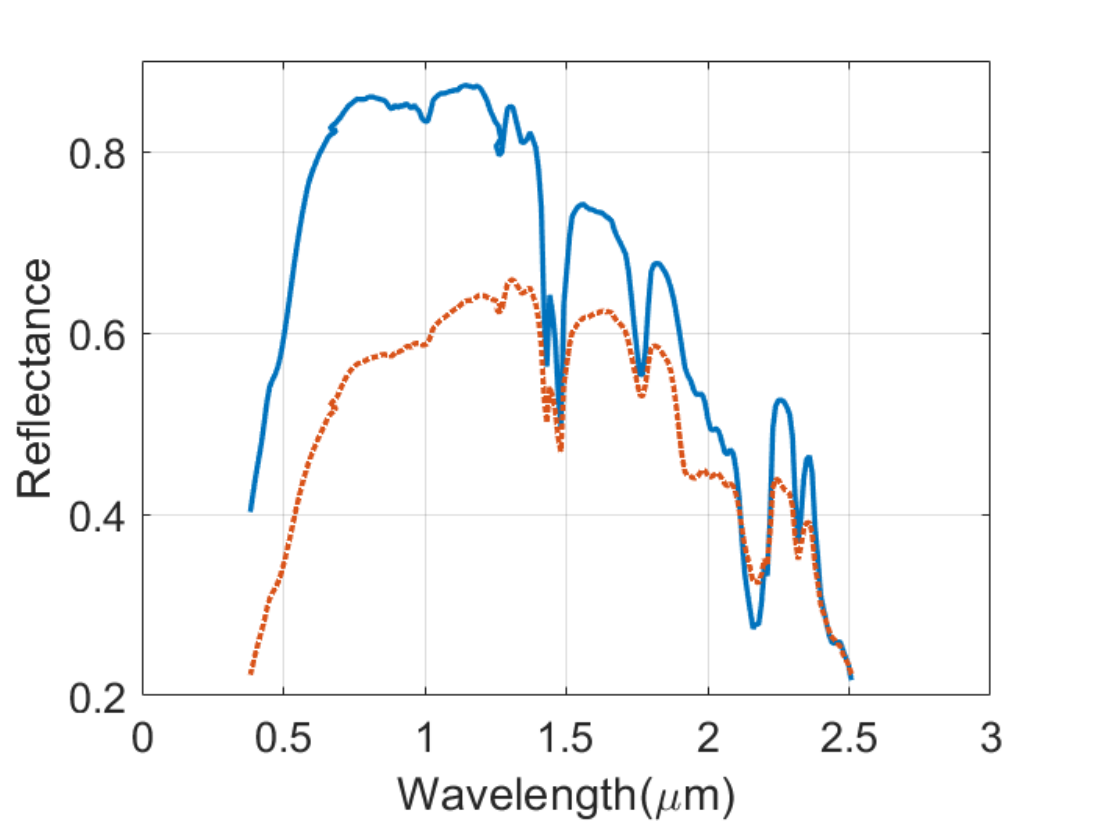}\\  (a) Chalcedony &(b) Alunite\\
\includegraphics[width=.4\linewidth]{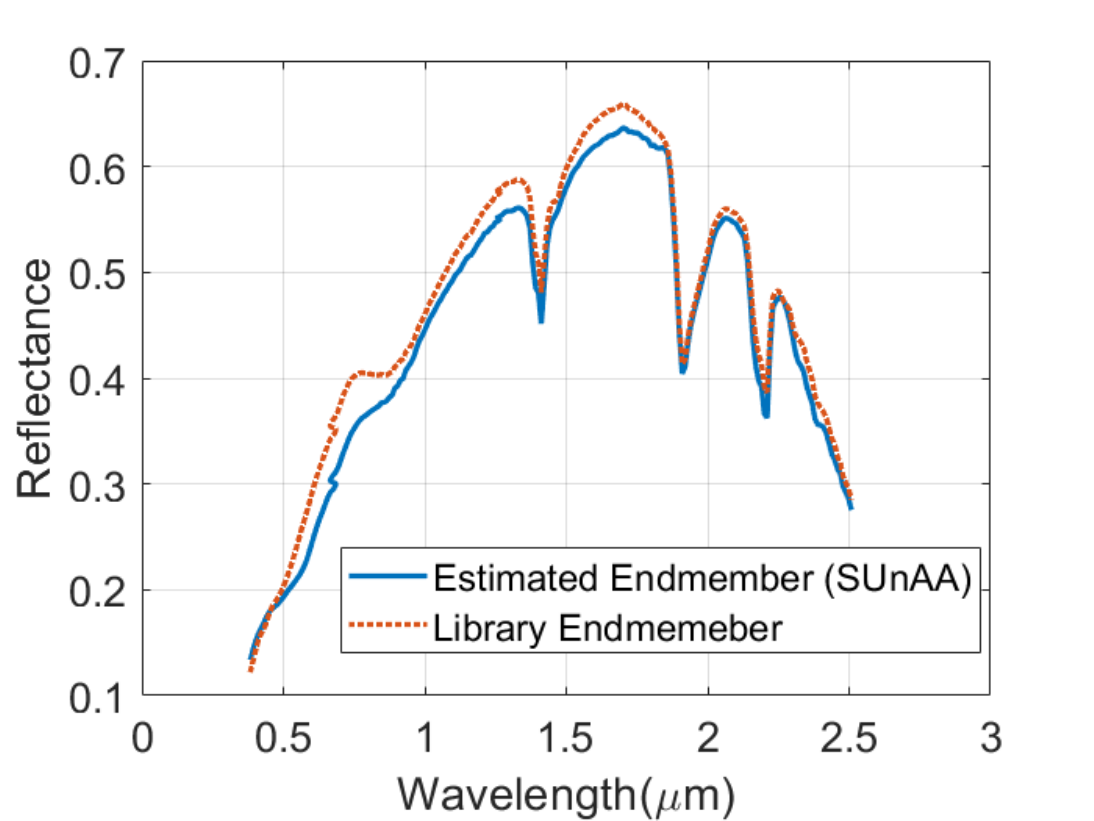}&\\  
(c) Kaolinite  &
 \end{tabular} 
\caption{Endmembers of three dominant minerals estimated using SUnAA applied to the Cuprite dataset compared with the ones from USGS library.}
\label{End_est}
\end{figure}

The substantial improvements of SUnAA can be attributed to the proposed model formulation leveraging archetypal analysis. Fig. \ref{End_est} depicts the estimated endmembers corresponding to the three minerals (Chalcedony, Alunite, and Kaolinite). Comparing the estimated endmembers with the corresponding ones from the library, \ie 297, 420 and 465, reveals that they are scaled versions. In real-world applications, the captured datasets are affected by noise, atmospheric effects, illumination variations, and the intrinsic variation of materials \cite{Spec_var}. Therefore, expecting the measured endmembers from a library to represent the materials in a real-world dataset is unrealistic. On the other hand, using archetypal analysis, we can achieve a linear combination of the endmembers from the dictionary to better represent the endmembers present in the scene.  

\section{Conclusion}
\label{sec: con}
We proposed a sparse unmixing technique using archetypal analysis called SUnAA. SUnAA models the endmembers of interest as a convex combination of endmembers from a library. We proposed a nonconvex optimization to simultaneously estimate the endmembers contributions and abundances. The proposed iterative algorithm is based on an active set method and is parameter-free. We evaluated SUnAA on two simulated and Cuprite datasets. The experimental results confirm that SUnAA leads to accurate abundances estimation and significantly outperforms the conventional sparse unmixing techniques. Additionally, experiments on a real-world dataset suggest that SUnAA can better detect existing minerals in a given scene. 

\section*{Acknowledgment}
The  work of  Alexandre Zouaoui was supported by ANR 3IA MIAI@Grenoble Alpes (ANR-19-P3IA-0003).

\ifCLASSOPTIONcaptionsoff
  \newpage
\fi

\bibliographystyle{IEEEbib.bst}
\bibliography{IEEEabrv,Ref}

\end{document}


%

\title{Supplement on: SUnAA: Sparse Unmixing using Archetypal Analysis}
%
%
%
\author{Behnood~Rasti*,~\IEEEmembership{Senior~Member,~IEEE,}
Alexandre~Zouaoui*,~\IEEEmembership{Student Member,~IEEE,} Julien~Mairal,~\IEEEmembership{Senior Member,~IEEE} and  Jocelyn Chanussot, ~\IEEEmembership{Fellow~Member,~IEEE} 
 \thanks{Behnood Rasti (corresponding author) is with Helmholtz-Zentrum Dresden-Rossendorf, Helmholtz Institute Freiberg for Resource Technology, Machine Learning Group, Chemnitzer Straße 40, 09599 Freiberg, Germany; b.rasti@hzdr.de, behnood.rasti@gmail.com}
\thanks{Alexandre Zouaoui, Jocelyn Chanussot, and Julien Mairal are with Univ. Grenoble Alpes, Inria, CNRS, Grenoble INP, LJK, 38000 Grenoble, France}
\thanks{* Behnood Rasti and Alexandre Zouaoui had an equal contribution.}
\thanks{Manuscript received .....}}

\maketitle

\section{archetypal analysis}
The archetypal analysis \cite{cutler1994archetypal}, which was also proposed for blind unmixing in \cite{EDAA}, assumes that the endmembers are convex combinations of the spectral pixels given by
\begin{argmini!}
  {\C, \A}{\|\Y - \Y \C \A\|_F^2,}{\label{eq: PLSU2}}{(\hat{\C}, \hat{\A}) = }
  \addConstraint{\C}{ \geq 0 \label{eq: ENC}}
  \addConstraint{{\bf \1}_n^T \C}{= {\1}_r^T \label{eq: ESC}}
  \addConstraint{\A}{ \geq 0 \label{eq: ANC}}
  \addConstraint{{\bf \1}_r^T \A}{= {\1}_n^T, \label{eq: ASC}}
\end{argmini!}
where $\C \in \mathbb{R}^{n\times r}$ corresponds to the contributions of the spectral pixels  $\Y$ and $\A \in \mathbb{R}^{r\times n}$ is the (low-rank) abundance matrix. 
\section{Experiments on Urban Dataset}

The Urban hyperspectral image contains
307x307 pixels captured by the Hyperspectral
Digital Image Collection Experiment (HYDICE) \cite{rickard_hydice_1993} sensor in 210 specrtral bands ranging from 400 to 2500
nm. After removing 48 noisy and water absorption bands 162 bands remain. The dataset have 4 endmembers, Asphalt Road, Grass, Tree, and Roof. The library ${\bf D}$ $ \in \mathbb{R}^{162\times 402}$ is composed of 402 spectral pixels obtained from the dataset.

\begin{figure*} [!t]\begin{center}
\begin{tabular}{c} 
\includegraphics[width=1\linewidth]{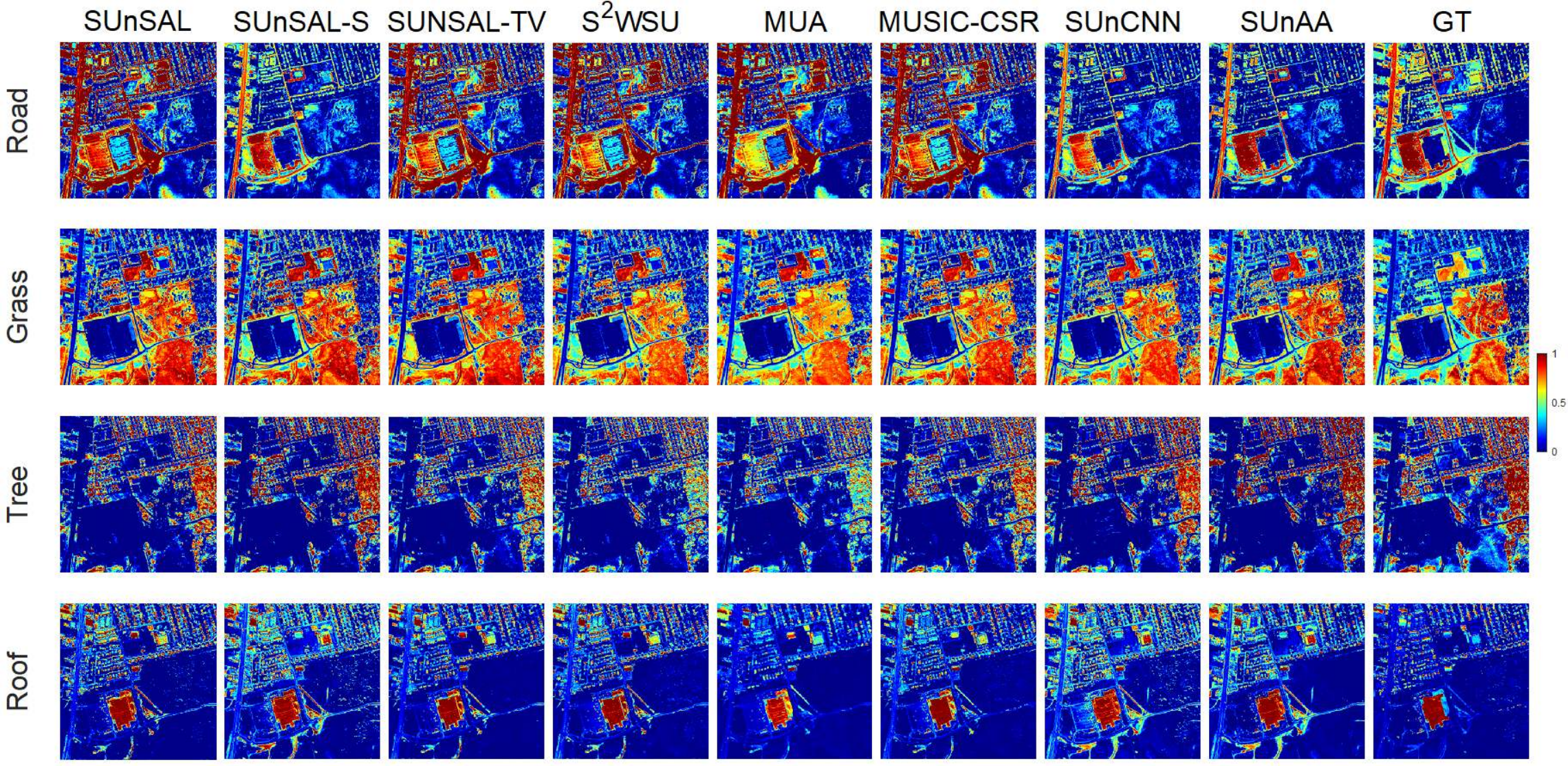}\\
 \end{tabular} \end{center} \caption{The fractional abundance of Urban datasets using different sparse unmixing techniques.}
 \label{Urabn}
\end{figure*}

Table \ref{tab:SRE_Urban} compares the sparse unmixing experiments applied to Urban in terms of SRE. As can be seen, SUnAA and SUnCNN perform similarly and outperform the other techniques. SUnSAL-S performs better than the other methods for this dataset. Fig. \ref{Urabn} visually compares the estimated abundances using different sparse unmixing techniques with the ground reference map.   

We should note that, in this experiment, we use $r=5$ for SUnAA. We notice that the library includes endmembers close to the zero endmember, often used for shadow removal, and SUnAA picks this endmember for $r=4$. Therefore, we select $r=5$, and since the ground truth abundances do not include the shadow abundance, we remove this from the abundances and normalize the abundances to be sum to one.
\begin{table}[htbp]\scriptsize
\centering\addtolength{\tabcolsep}{-5.5pt}
\caption{Sparse unmixing experiments applied to Urban in terms of SRE. The best performances are shown in bold. The second best are underlined.}
\begin{tabular}{c|c|c|c|c|c|c|c|c}
		\toprule
		SNR    &SUnSAL   & SUnSAL-S & SUnSAL-TV             &S$^2$WSU & MUA & MUSIC-CSR & SUnCNN & SUnAA \\		
		\midrule
		Urban             & 1.73 & 7.10 & 2.07 & 1.58 & 2.65 & 2.24 & \underline{7.74} & \textbf{7.77} \\
		\bottomrule
	\end{tabular}        
	\label{tab:SRE_Urban}
\end{table}

\section{Processing time}

Table \ref{tab:time} gives the processing times for different unmixing techniques applied to Urban and Cuprite datasets. SUnSAL, SUnSAL-S, SUnSAL-TV, S$^2$WSU,  and MUA were implemented in Matlab (2020b) and MUSIC-CSR, SUnCNN, and SUnAA were implemented in Python (3.8). The reported processing times were obtained using a computer with Intel(R) Xeon(R) Silver 4112 CPU  2.60GHz, eight cores, 64GB of memory, and an NVIDIA TITAN RTX with 24GB of memory.  The table shows that MUA is the most and SUnSAL-TV is the least efficient method for both datasets. SUnAA is competitive to MUA in the case of Urban and outperforms the other techniques. However, in the case of Cuprite, SUnAA is not as efficient. We should note that the increase in the processing time of SUnAA is due to the column-wise update of matrix ${\bf B}$. Therefore, the higher number of endmembers will decrease the algorithm's speed.

\begin{table}[htbp]\scriptsize
\centering\addtolength{\tabcolsep}{-5.5pt}
\caption{Processing times of different sparse unmixing techniques applied to real datasets.}
\begin{tabular}{c|c|c|c|c|c|c|c|c}
		\toprule
		HS    &SUnSAL   & SUnSAL-S & SUnSAL-TV             &S$^2$WSU & MUA & MUSIC-CSR & SUnCNN & SUnAA \\		
		\midrule
			Urban  & 672.5 & 1550.4 & 2107.9 & 606.5 & \textbf{216.0} & 1975.1 & 320.8 & \underline{306.2}\\ \midrule
		Cuprite & 224.0 & \underline{203.4} & 1109.2 & 389.8 & \textbf{160.8} & 318.4 & 904.8 & 703.5\\
		\bottomrule
	\end{tabular}       
	\label{tab:time}
\end{table}

%


\ifCLASSOPTIONcaptionsoff
  \newpage
\fi

\bibliographystyle{IEEEtran}
\bibliography{Ref}
